\documentclass[11pt]{article}
\usepackage[hyperref]{ccl2024-en}
\usepackage{times}
\usepackage{url}
\usepackage{latexsym}
\usepackage{fancyhdr}

\usepackage{amsfonts}
\usepackage{amsmath}
\usepackage{subcaption}
\usepackage{tikz}
\usepackage{pgfplots}
\usepackage{multirow}
\usepackage{booktabs}
\usepackage{array}
\usepackage{pgfplotstable}
\pgfplotsset{compat=1.18}
\definecolor{goodforeyesgreen}{RGB}{67,151,143}
\definecolor{lovelyred}{RGB}{229,111,94}
\definecolor{sadblue}{RGB}{30,144,255}
\definecolor{supriseorange}{RGB}{255,183,127}

\newcommand{\confidenceband}[5][]{ 
\pgfplotstableread{#2}\datatable
    \addplot [draw=none, stack plots=y, forget plot] table [
        x={#3},
        y expr=\thisrow{#5}
    ] {\datatable};
    \addplot [draw=none, fill=gray!40, stack plots=y, area legend, #1] table [
        x={#3},
				y expr=\thisrow{#4}-\thisrow{#5}
    ] {\datatable} \closedcycle;
    \addplot [forget plot, stack plots=y,draw=none] table [x={#3}, y expr=-\thisrow{#4}] {\datatable};
}

\title{Prior Constraints-based Reward Model Training for Aligning Large Language Models}

\author{
    Hang Zhou$^{1}$,
    Chenglong Wang$^{1}$,
    Yimin Hu$^{1}$,
    Tong Xiao$^{1,2}$\thanks{$\ $ Corresponding author.}, \\ 
    \textbf{Chunliang Zhang$^{1}$,
    and Jingbo Zhu$^{1,2}$} \\
	$^{1}$NLP Lab, School of Computer Science and Engineering, \\
    Northeastern University, Shenyang, China \\
	$^{2}$NiuTrans Research, Shenyang, China \\
	\tt \{stceum, clwang1119\}@gmail.com,\\
    \tt \{xiaotong, zhujingbo\}@mail.neu.edu.cn
}

\date{}

\begin{document}
\maketitle
\begin{abstract}

Reinforcement learning with human feedback for aligning large language models (LLMs) trains a reward model typically using ranking loss with comparison pairs. 
However, the training procedure suffers from an inherent problem: the uncontrolled scaling of reward scores during reinforcement learning due to the lack of constraints while training the reward model.
This paper proposes a \textbf{P}rior \textbf{C}onstraints-based \textbf{R}eward \textbf{M}odel (\textbf{PCRM}) training method to mitigate this problem.
PCRM incorporates prior constraints—specifically, length ratio and cosine similarity between outputs of each comparison pair—during reward model training to regulate optimization magnitude and control score margins.
We comprehensively evaluate PCRM by examining its rank correlation with human preferences and its effectiveness in aligning LLMs via RL. Experimental results demonstrate that PCRM significantly improves alignment performance by effectively constraining reward score scaling.
As another bonus, our method is easily integrated into arbitrary rank-based alignment methods, such as direct preference optimization, and can yield consistent improvement.
The code is available at \url{https://github.com/wangclnlp/DeepSpeed-Chat-Extension/tree/PCRM}. 

\end{abstract}

\section{Introduction}
\label{sec:intro}

%
%

\cclfootnote{
    %
    %
    \hspace{-0.65cm}  
    \textcopyright 2024 China National Conference on Computational Linguistics

    \noindent Published under Creative Commons Attribution 4.0 International License
}

Reinforcement learning with human feedback (RLHF) has been proven to be an advanced technology to align large language models (LLMs) with human preferences \cite{ouyang2022training,ji2023ai,wang2023esrl}. 
It builds upon preference data, which rates and compares different outputs given the same input, where this rating is conducted by either human annotators or LLMs \cite{ouyang2022training,lee2023rlaif,cui2023ultrafeedback,dubois2024alpacafarm}.
In practice, RLHF trains a reward model on the preference data with ranking loss for higher scores on preferred outputs than dispreferred ones.
Then, RL algorithms such as Proximal Policy Optimization (PPO) \cite{schulman2017proximal} are employed to fine-tune the LLM with the aim of optimizing this reward.
During RL training, the reward model will give scores as signals, and the LLM will align the preference by increasing the probabilities of sampled outputs with higher reward scores.
It is discernible that the alignment of the LLM is significantly influenced by how well the reward model is trained.

However, the training procedure of the reward model actually suffers from a failure mode: it learns from the preference data with the standard ranking loss yet provides a reward score for the sampled outputs during RL training. 
This training mode causes an inherent problem: the ranking loss increases the score margin between the outputs of this comparison without constraint during the training procedure, which makes an uncontrollable scale of scores in the process of RL training.
For instance, the reward model trained by ranking loss can predict reward scores with a relatively large margin to two sampled outputs that do not differ very much \cite{zhu2023principled}.

To address the problem, we propose a \textbf{P}rior \textbf{C}onstraints-based \textbf{R}eward \textbf{M}odel (\textbf{PCRM}) training method in this paper, which incorporates prior constraints while training the reward model. 
Specifically, we select two features of the preference data for designing prior constraints: length ratio and cosine similarity of the outputs with the same prompt. 
The length ratio is computationally less demanding and directly reflects data differences, while the cosine similarity captures deeper semantic features despite being more computationally intensive. 
These constraints effectively regulate the optimization magnitude across different outputs for the same input, thereby controlling the margin of scores predicted by the reward model.

We comprehensively evaluate the proposed PCRM in the following two ways.
Firstly, we test the rank correlation between the preferences predicted by PCRM and human preferences.
It can assess the extent to which the reward model can serve as a surrogate for human-derived preference signals.
Secondly, we verify the effectiveness of the PCRM in aligning an LLM via RL.
This also demonstrates the influence of constraints on alignment performance straightly.
Notably, PCRM can yield a +2.48\% improvement in the GPT-4 win rate for the dialogue task compared to the traditional RLHF.
Furthermore, we integrate our method into direct preference optimization (DPO), a rank-based alignment method.
The results show that our method can also be effective in improving the rank-based alignment methods, \textit{e.g.,} a 2.95\% increase in the GPT-4 win rate on the dialogue task compared to DPO.

\section{Related Work}

Reinforcement learning with human feedback (RLHF) is a crucial technique to ensure that the behaviours of large language models (LLMs) are consistent with human preferences \cite{stiennon2020learning,ouyang2022training,wang2023esrl}.
Recent works resorted to building a better reward model to improve the performance of RLHF on LLMs.
These methods could be classified into three groups.
The first group aimed to efficiently produce human preference data \cite{dubois2024alpacafarm,cui2023ultrafeedback,lee2023rlaif}.
For instance, \newcite{lee2023rlaif} directly employed an LLM to annotate the comparison pairs.
Significantly, the reward model trained on the LLM-annotated comparison pairs can achieve closer performance when applied to RLHF than the one trained on human-annotated comparison pairs.
The second group tended to design fine-grained reward models to provide multiple reward scores for different reward criteria \cite{cheng2023everyone,wang2023esrl,wu2024fine,zhong2024panacea}.
Notably, \newcite{wang2023esrl} learned various evaluation models from an LLM as reward models on the summarization task, including reward models scored for relevance, scored for fluency, scored for consistency, and scored for coherence.
The third group that has attracted less attention generally explored how to merge multiple reward models in RLHF, such as reward model ensemble \cite{coste2023reward} and learning reward weights \cite{min2024dynamic}.
Different from these methods, in this paper, we proposed a superior training schema that enables this reward model to give a more accurate score in RLHF.
Specifically, we refined the conventional ranking loss, typically utilized in reward model training, by incorporating prior constraints. These constraints are designed to regulate the margin between the scores of various outputs generated from the same input.

To the best of our knowledge, there is almost no previous work in training a reward model with prior constraints.
The few related one is LLaMA2 \cite{touvron2023llama}, which added a margin component in ranking loss for training each comparison pair.
Although this margin simulates a constraint, it requires a manual annotation, which significantly increases the cost of training the reward model.
Additionally, this work did not provide a specialized exploration or analysis on the topic of constraining the margin of scores during training reward models.
In comparison, all of our prior constraints were computed automatically and did not require any human annotation.
We also provided sufficient theoretical analysis and experiments to constrain the margin of scores.
Furthermore, although there are some methods such as DPO \cite{amini2024direct} and RRHF \cite{yuan2023rrhf} that circumvent the need for training reward models, they still suffer from an inherent limitation from ranking loss, which is mentioned in DPO \cite{amini2024direct} without a solution.
Our proposed PCRM can be easily extended to these methods to yield some benefits (See Section \ref{sec:extend_dpo}).

\section{Background}
Human-preference alignment training is a key technique to ensure that the behaviours of LLMs are consistent with human preferences.
Recent efforts to align LLMs have mainly been conducted via RLHF.
It typically includes three stages: 1) collecting preference data, 2) training a reward model with ranking loss, and 3) optimizing an LLM against the reward model via RL.

\subsection{Collecting Preference Data}
The preference data consists of the given input $x$ and the corresponding different outputs sampled from LLMs trained by SFT (denoted as $\pi^{\textrm{SFT}}_{\theta}$) or annotated by humans, which we refer to as $(y_1, y_2, \cdots, y_n|x)$.
In the preference data, the different outputs are rated and ranked by humans or LLMs with specific aspects \cite{ouyang2022training,dubois2024alpacafarm}, which are denoted as $(y^{(1)} \succ y^{(2)} \succ \cdots \succ y^{(i)} \succ y^{(j)} \succ \cdots \succ y^{(n)}|x)$, where $y^{(1)}$ is the best while $y^{(n)}$ is the worst.


\subsection{Training Reward Model}
After preference data collection, we can train a reward model $\pi^{\textrm{RM}}_{\theta}$ from the preference data, where the training objective is to fit human preference.
Note that we use $\pi^{\textrm{SFT}}_{\theta}$ to initialize the reward model.
Suppose $r^*$ as the ideal model of human preference, based on the Bradley-Terry model \cite{bradley1952rank}, the distribution of human preference $P_{\textrm{Human}}\left(y^{(i)} \succ y^{(j)}|x\right)$ can be written as:
\begin{eqnarray}
    P_{\textrm{Human}}\left(y^{(i)} \succ y^{(j)}|x\right)
        &=& \frac{\exp\left({r^*\left(y^{(i)},x\right)}\right)}
                {\exp\left({r^*(y^{(i)},x)}\right)+\exp\left({r^*(y^{(j)},x)}\right)} \\
        &=& \sigma\left(r^*(y^{(i)},x)-r^*(y^{(j)},x)\right)
\end{eqnarray}
where $y^{(i)}$ denotes the data ranking $i$ and $\sigma$ denotes the Sigmoid activation function. 
When dealing with multiple outputs more than two, similarly, we can induce $P_{\textrm{Human}}(\cdot)$ based on the more general Plackett-Luce model \cite{plackett1975analysis,luce2005individual}:
\begin{eqnarray}
    P_{\textrm{Human}}\left(y^{(1)} \succ y^{(2)} \succ \cdots \succ y^{(i)} \succ y^{(j)}
        \succ \cdots \succ y^{(n)}|x\right)
        = \prod_{i=1}^{n} \frac{\exp\left({r^*(y^{(i)},x)}\right)}
                                {\sum_{j=i}^{n} \exp\left({r^*(y^{(j)},x)}\right)}
\end{eqnarray}
To learn the preference distribution, we need to increase the probability of preferred outputs.
Here, it is typically achieved by a negative log-likelihood loss function:
\begin{eqnarray}
    \mathcal{L}_{r} &=& -\log P_{\textrm{Human}}\left(y^{(1)} \succ y^{(2)} \succ \cdots \succ 
                                y^{(i)} \succ y^{(j)} \succ \cdots \succ y^{(n)}|x\right) \\
                    &=& - \sum_{i=1}^{n} \log \frac{\exp\left({r^*\left(y^{(i)},x\right)}\right)}
                                                {\sum_{j=i}^{n} \exp\left({r^*\left(y^{(j)},x\right)}\right)}
\end{eqnarray}
Specially, when $n=2$ which means there is only a comparison pair, the loss would be:
\begin{eqnarray}
    \mathcal{L}_{r} &=& -\log P_{\textrm{Human}}\left(y^{(i)} \succ y^{(j)}\right) \\
                    &=& -\log \sigma \left(r^*\left(y^{(i)},x\right)-r^*\left(y^{(j)},x\right)\right)
\end{eqnarray}

\subsection{RL Training Against the Reward Model}
In the process of RL training, we use the reward output $\pi^{\textrm{RM}}_{\theta}$ as signals, combined with an RL algorithm.
Taking PPO as an instance, the corresponding loss for this training sample is given by: 
\begin{eqnarray}
\max_{\pi_\theta}
\mathbb{E}_{x\sim\mathcal{D},y\sim\pi_\theta}
\left[\pi^{\textrm{RM}}_{\theta}(x,y)\right]
-\beta\mathbb{D}_\mathrm{KL}\left[\pi_\theta||\pi^{\textrm{SFT}}_{\theta}\right]
\end{eqnarray}
where $\mathcal{D}$ is the dataset of RL training, $x$ is the input, and $y$ for the sampled outputs.
$\mathbb{D}_\mathrm{KL}$ is the KL dispersion which measures the distributional difference between $\pi^{\textrm{SFT}}_{\theta}$ and $\pi_\theta$, multiplied by $\beta$ which controls their distance, as the bigger $\beta$ is, the more significant constraint is applied.

\section{Prior Constraints-based Reward Model}
Motivated by the challenges posed by the uncontrollable scale of scores in the reward model during the RL training process, our aim is to constrain the score margin between the outputs of this comparison when training the reward model.
We propose the PCRM method to achieve this.
Unlike conventional ranking loss, the proposed PCRM can constrain the maximum score margin between the outputs of each comparison with the length and cosine similarity features.
In the following subsections, we will describe our PCRM in detail. 

\subsection{Optimization Objective of PCRM} \label{sec:training_object_of_PCRM}
Given the input $x$ and the reward model $\pi^{\textrm{RM}}_\theta$, the probability of $y_1 \succ y_2$ can be written as:
\begin{eqnarray}
    &&P_{\pi^{\textrm{RM}}_\theta}\left(y_1 \succ y_2|x\right) \label{eq:vanilla_rm_prob} \\ 
    &=& P_{\pi^{\textrm{RM}}_\theta}(y_1-y_2\succ0|x) \\
    &=& \sigma\left(\pi^{\textrm{RM}}_\theta(y_1,x)-\pi^{\textrm{RM}}_\theta(y_2,x)\right) \label{eq:sigmoid_pi1-pi2}\\
    &=& \sigma\left(\Delta_{\pi^{\textrm{RM}}_\theta}(y_1, y_2, x)\right) \label{eq:vanilla_reward_prob}
\end{eqnarray}
where $\Delta_{\pi^{\textrm{RM}}_\theta}(\cdot)$ denotes the margin in the evaluation scores predicted by the reward model $\pi^{\textrm{RM}}_\theta$.
If the reward model learns this probability with a standard ranking loss, the reward score of the preferred output will increase; conversely, the reward score of the dispreferred output will decrease.
This endeavour aims to maximize the margin between the score of the preferred and dispreferred outputs as much as possible.
However, in RLHF, we do not just expect the reward model to be able to distinguish which output is more preferred, but also to be able to give information on \textit{how much} more one input is preferred.
Consequently, we conjecture that facilitating the reward model to learn an appropriate score margin across different outputs could improve the performance of RLHF.

We achieve this goal by adding a maximum margin constraint, denoted as $\Delta^*(\cdot)$, where it takes values in the range $(0, +\infty)$.
We re-derive Equation \ref{eq:vanilla_rm_prob} with the constraint $\Delta^*(\cdot)$ as follows:
\begin{eqnarray}
    &&P_{\pi^{\textrm{RM}}_\theta}\left(\Delta^*(y_1,y_2,x) \succ y_1 - y_2 \succ 0|x\right) \\ 
    &=&P_{\pi^{\textrm{RM}}_\theta}\left(\Delta^* > \Delta_{\pi^{\textrm{RM}}_\theta}>0\right) \\ 
    &=& P_{\pi^{\textrm{RM}}_\theta}\left(\Delta^*> \Delta_{\pi^{\textrm{RM}}_\theta}\right) \times 
        P_{\pi^{\textrm{RM}}_\theta}\left(\Delta_{\pi^{\textrm{RM}}_\theta}>0\right) \\
    &=& \sigma\left(\Delta^* - \Delta_{\pi^{\textrm{RM}}_\theta}\right) \times \sigma\left(\Delta_{\pi^{\textrm{RM}}_\theta}\right) \label{eq:PCRM_prob}
\end{eqnarray}
Based on the above derivation, we have the negative log-likelihood loss similar to the vanilla one:
\begin{eqnarray}
    \mathcal{L}_{\textrm{PCRM}} &=& -\log P_{\pi^{\textrm{RM}}_\theta}\left(\Delta^* > \Delta_{\pi^{\textrm{RM}}_\theta}>0\right) \\
                    &=& -\log \sigma \left(\Delta^* - \Delta_{\pi^{\textrm{RM}}_\theta}\right) -\log \sigma \left(\Delta_{\pi^{\textrm{RM}}_\theta}\right) \label{eq:PCRM_loss}
\end{eqnarray}

We define the maximum margin constraint through a negative correlation with the similarity between $y_1$ and $y_2$:
\begin{eqnarray}
    \label{eq:constraint_similarity}
    \Delta^*(y_1, y_2, x) = \frac{\beta_1}{\textrm{Sim}(y_1,y_2,x)+\beta_2}+\beta_3
\end{eqnarray}
where $\textrm{Sim}(\cdot)$ denotes the similarity of the different outputs.
$\beta_1$ controls the magnitude of the $\Delta^*(\cdot)$, $\beta_2$ controls the range of variation and $\beta_3$ controls the offset. 
We employ length ratio and cosine similarity to estimate $\textrm{Sim}(\cdot)$.
When using the length ratio to estimate $\textrm{Sim}(\cdot)$, $\textrm{Sim}(\cdot)$ can given by:
\begin{eqnarray}
    \textrm{Sim}_{\textrm{len\_rat}}(y_1,y_2,x) &=& \frac{\min(\phi(y_1),\phi(y_2))}{\max(\phi(y_1),\phi(y_2))}
\end{eqnarray}
where $\phi(y)$ denotes the length of $y$. 
Note that we do not need the input $x$ to be involved when employing the length ratio to estimate the $\textrm{Sim}(\cdot)$.
When considering the cosine similarity as another estimation of the $\textrm{Sim}(\cdot)$, written as:
\begin{eqnarray}
    \textrm{Sim}_{\textrm{cos\_sim}}(y_1,y_2,x) 
    &=& 1 - \frac{1}{\pi}\times\arccos\left(\frac{E(x,y_1) \cdot E(x,y_2)}{\max(||E(x,y_1)||_2, \epsilon) \cdot \max(|| E(x,y_2) ||_2, \epsilon)}\right) \label{eq:sim_cos_sim}
\end{eqnarray}
where $\epsilon$ is a small value to avoid division by zero and $\arccos(\cdot)$ is the inverse of the cosine function.
Inspired by the strong text encoding capability of the pre-trained model \cite{devlin2018bert,xiao2023introduction}, we employ a pre-trained model like BERT to compute $E(\cdot)$.
There are other choices to define $\textrm{Sim}(\cdot)$ for specific tasks.
For instance, we can use the ROUGE function \cite{lin2004rouge} to define $\textrm{Sim}(\cdot)$ in the summarization task. 

\subsection{Analysis of the Optimization for PCRM} 
To better understand the constrained optimization for PCRM, we take the derivative of the loss function. The gradient of $\mathcal{L}_{\textrm{PCRM}}$ with respect of the parameters $\theta$ is:
\begin{eqnarray}
\nabla_\theta\mathcal{L}_{\textrm{PCRM}}
=\nabla_\theta\Delta_{\pi^{\textrm{RM}}_\theta}\times\left(\sigma\left(\Delta_{\pi^{\textrm{RM}}_\theta}-\Delta^*\right)-\sigma\left(-\Delta_{\pi^{\textrm{RM}}_\theta}\right)\right) \label{eq:grad_pcrm}
\end{eqnarray}
As a comparison, we also take the derivative of one for vanilla reward loss, which can be written as:
\begin{eqnarray}
\nabla_\theta\mathcal{L}_{\textrm{RM}}
=\nabla_\theta\Delta_{\pi^{\textrm{RM}}_\theta}\times\left(-\sigma\left(-\Delta_{\pi^{\textrm{RM}}_\theta}\right)\right) \label{eq:grad_vanilla_rm}
\end{eqnarray}
Compare Equation \ref{eq:grad_pcrm} with \ref{eq:grad_vanilla_rm}, we can find their difference $\sigma\left(\Delta_{\pi^{\textrm{RM}}_\theta}-\Delta^*\right)$ which is always positive and will decrease the coefficient of $\nabla_\theta\Delta_{\pi^{\textrm{RM}}_\theta}$, constraining the optimization thereby.

When $\sigma\left(\Delta_{\pi^{\textrm{RM}}_\theta}-\Delta^*\right)-\sigma\left(-\Delta_{\pi^{\textrm{RM}}_\theta}\right)=0$ in Equation \ref{eq:grad_pcrm}, it implies that $\Delta_{\pi^{\textrm{RM}}_\theta}-\Delta^*=-\Delta_{\pi^{\textrm{RM}}_\theta}$ due to the monotonic increase of the Sigmoid activation function. 
Thus, we can deduce $\Delta_{\pi^{\textrm{RM}}_\theta}=\frac{\Delta^*}{2}$.
Then we can get the conclusion that when $\Delta_{\pi^{\textrm{RM}}_\theta}<\frac{\Delta^*}{2}$, the coefficient of $\nabla_\theta\Delta_{\pi^{\textrm{RM}}_\theta}$ has the same sign with the one of the vanilla reward gradient, meaning the same optimization direction with the origin; in contrast, when $\Delta_{\pi^{\textrm{RM}}_\theta}>\frac{\Delta^*}{2}$, there will be opposite optimization direction decreasing the margin of reward scores.
In this way, we can control the distance of scores of different outputs while optimizing the reward model.



\section{Experiments}

We evaluate the proposed PCRM in the following two ways.
Firstly, we test the reward model trained by PCRM.
Secondly, we analyze the performance of applying this trained reward model to RLHF.
We conduct experiments on the commonly used generation tasks, including dialogue and summarization.

\subsection{Datasets}
The datasets used for each task are as follows:
\begin{itemize}
    \item \textit{Dialogue}: We employed AlpacaFarm \cite{dubois2024alpacafarm} dataset on dialogue task, which consists of 10K supervised fine-tuning split, 10K pairwise preference split, 20K unlabeled split for RL training, and 2K validation split based on 52k Alpaca data \cite{taori2023alpaca}.
    For evaluation, we employed their evaluation set, which contains 805 instructions selected from a series of open-source datasets with real-world user interactions as reference instructions.

    \item \textit{Summarization}: We used the filtered versions of the TL;DR dataset and human feedback dataset provided by OpenAI \cite{stiennon2020learning} on summarization task, the former one for instruction tuning and alignment, and the latter one for reward modeling. 
    The TL;DR dataset is filtered to ensure quality and contains 123.2K samples in the final version, including 116.7K for training, 6.4K for validating, and 6.5K for testing. 
    The large, high-quality preference dataset of human comparisons between summaries contains a 92.9K training set, a 33.1K validation set, and a 50.7K test set. 
    Due to the enormous computational cost caused by the vast testing set, we randomly selected 10\% as our final test set. 
\end{itemize}

\subsection{Settings}

\begin{table}[!ht]
    \centering
    \begin{tabular}{llccrc}
    \toprule
    \textbf{Task} & $\textbf{\textrm{Sim}}(\cdot)$ & $\boldsymbol\beta_1$ & $\boldsymbol\beta_2$ & $\boldsymbol\beta_3$ & \textbf{max\_len} \\
    \midrule
    \multirow{2}{*}{Dialogue} & len\_rat & 10 & 0.001 & -5 & \multirow{2}{*}{512} \\
    & cos\_sim & 20 & 0.001 & -15 & \\
    \midrule
    \multirow{2}{*}{Summarization} & len\_rat & 10 & 0.001 & -5 & \multirow{2}{*}{1024} \\
    & cos\_sim & 20 & 0.001 & -15 & \\
    \bottomrule
\end{tabular}
    \caption{The hyper-parameters used in our experiments.}
    \label{tab:hyperparameters}
\end{table}

We conducted our experiments based on DeepSpeed-Chat\footnote[1]{\url{https://github.com/microsoft/DeepSpeedExamples}} with a cross-entropy loss on supervised fine-tuning, where prompt parts were masked, and a ranking loss was employed by reward modeling.
We set the maximum sequence length to 512 and 1024 for the dialogue and summarization tasks, respectively.
For calculating the similarity between pair-wise data for setting prior constraints, we employed the BERT-base model \cite{devlin2018bert} to encode paired data and then calculated the similarity between \texttt{[CLS]} embeddings.
Considering that the maximum sequence length supported by BERT is 512; however, the sequence length is 1024 for summarization, we set the tokenizer to truncate from the left, \textit{i.e.,} truncating part of the prompt and calculating the semantic similarity of the remaining part.
Additionally, the value of $\beta_1$, $\beta_2$ and $\beta_3$ mentioned in Eq. \ref{eq:constraint_similarity} are shown in Table \ref{tab:hyperparameters}.
The performance of other hyper-parameters is reported in Section \ref{subsubsec:value_constraints_on_PCRM}.
Furthermore, we employed the top-$p$ sampling method in the process of generation, where the temperature and the $p$ were set to 0.75 and 0.95, respectively.

\subsection{Evaluation Metrics}

We evaluated the effectiveness of our method comprehensively from two dimensions: training reward models and aligning LLMs with the trained reward models.
For training reward models, we scored the pair-wise test set.
Based on this, we calculated the accuracy of the predicted scores (for details of calculating accuracy, see Appendix A).
We trained three epochs on the training set and saved the model for each epoch.
We selected the best one with the validation set. 
We compared the performance of the vanilla method with ours, as shown in the following section.
To align LLMs with the trained reward models, we used different metrics depending on the tasks. 
Specifically, for the dialogue task, we used PandaLM \cite{wang2023pandalm} and GPT-4 to choose a better one from the model output and the reference and calculate the win rate following \newcite{rafailov2024direct}.
For the summarization task, except for GPT-4, we also employ ROUGE \cite{lin2004rouge} and BARTScore \cite{yuan2021bartscore} to evaluate the quality of the summary generated by the model from the posts.

\subsection{Results of Training Reward Models}

\begin{table}[!t]
    \centering
    \begin{tabular}{l>{\centering\arraybackslash}p{8em}>{\centering\arraybackslash}p{8em}}
    \toprule
    \textbf{Method} & \textbf{Dialogue} & \textbf{Summarization} \\
    \midrule
    Vanilla RM & 54.93 & 72.20 \\
    PCRM-Random & 54.13 & 71.70 \\
    PCRM-Fixed & 54.40 & 71.70 \\
    PCRM-Length (Ours) & 55.87 & 72.50 \\
    PCRM-Cosine (Ours) & \textbf{56.53} & \textbf{72.60} \\
    \bottomrule
\end{tabular}
    \caption{The accuracy of reward models on the dialogue and summarization task. 
    ``-Random'' means using random constraints, ``-Fixed'' means using fixed one, ``-Cosine'' means using constraints calculated by cosine similarity, and ``-Length'' for length ratio.}
    \label{tab:reward_accuracy}
\end{table}

\begin{figure}[!t]
    \centering
    \begin{subfigure}[b]{.45\textwidth}
	\centering
	\begin{tikzpicture}[domain=0:1, scale=0.8]
		\begin{axis}[xlabel={$\textrm{Sim}_{\textrm{cos\_sim}}$}, ylabel={\normalsize{Margin of Reward Scores ($\Delta_{\pi_{\textrm{RM}}}$})},
			y label style={yshift=-.5em}, x label style={yshift=1ex}, enlarge x limits=0.02, enlarge y limits=0.02, xmin=0.9, ymax=.45]
			\addplot [scatter, only marks,  scatter src=explicit symbolic,
				scatter/classes={
					r={mark=*,goodforeyesgreen, fill opacity=0.2, draw opacity=0},
					w={mark=diamond*,draw=black!30,fill=black!30,fill opacity=0.8, draw opacity=0}
			}]
			table [x=sim, y=delta,meta=label]{"figures/data/alpaca/baseline-epoch1-delta-sim_cos_sim.dat"};
		\end{axis}
	\end{tikzpicture}
\end{subfigure}
\begin{subfigure}[b]{.45\textwidth}
	\centering
	\begin{tikzpicture}[domain=0:1,scale=0.8]
		\begin{axis}[xlabel={$\textrm{Sim}_{\textrm{len\_rat}}$}, ylabel={\normalsize{Margin of Reward Scores ($\Delta_{\pi_{\textrm{RM}}}$})},
			y label style={yshift=-.5em}, x label style={yshift=1ex}, enlarge x limits=0.02, enlarge y limits=0.02, xmin=0.4, ymax=.45]
			\addplot [scatter, only marks,  scatter src=explicit symbolic,
				scatter/classes={
					r={mark=*,goodforeyesgreen, fill opacity=0.2, draw opacity=0},
					w={mark=diamond*,draw=black!30,fill=black!30,fill opacity=0.8, draw opacity=0}
			}]
			table [x=sim, y=delta,meta=label]{"figures/data/alpaca/baseline-epoch1-delta-sim_len_rat.dat"};
		\end{axis}
	\end{tikzpicture}
\end{subfigure} \\
\normalsize{(a) w/o constraints} \\
\vspace{0.2cm}
\begin{subfigure}[b]{.45\textwidth}
	\centering
	\begin{tikzpicture}[domain=0:1,scale=0.8]
		\begin{axis}[xlabel={$\textrm{Sim}_{\textrm{cos\_sim}}$}, ylabel={\normalsize{Margin of Reward Scores ($\Delta_{\pi_{\textrm{RM}}}$})},
			y label style={yshift=-.5em}, x label style={yshift=1ex}, enlarge x limits=0.02, enlarge y limits=0.02, xmin=0.9, ymax=.45]
			\addplot [scatter, only marks,  scatter src=explicit symbolic,
				scatter/classes={
					r={mark=*,goodforeyesgreen, fill opacity=0.2, draw opacity=0},
					w={mark=diamond*,draw=black!30,fill=black!30,fill opacity=0.8, draw opacity=0}
			}]
			table [x=sim, y=delta,meta=label]{"figures/data/alpaca/cos_sim-beta1_20_beta3_-15-delta-sim_cos_sim.dat"};
		\end{axis}
	\end{tikzpicture}
\end{subfigure}
\begin{subfigure}[b]{.45\textwidth}
	\centering
	\begin{tikzpicture}[domain=0:1,scale=0.8]
		\begin{axis}[xlabel={$\textrm{Sim}_{\textrm{len\_rat}}$}, ylabel={\normalsize{Margin of Reward Scores ($\Delta_{\pi_{\textrm{RM}}}$})},
			y label style={yshift=-.5em}, x label style={yshift=1ex}, enlarge x limits=0.02, enlarge y limits=0.02, xmin=0.4, ymax=.45]
			\addplot [scatter, only marks,  scatter src=explicit symbolic,
				scatter/classes={
					r={mark=*,goodforeyesgreen, fill opacity=0.2, draw opacity=0},
					w={mark=diamond*,draw=black!30,fill=black!30,fill opacity=0.8, draw opacity=0}
			}]
			table [x=sim, y=delta,meta=label]{"figures/data/alpaca/len_rat-beta1_10_beta3_-5-delta-sim_len_rat.dat"};
		\end{axis}
	\end{tikzpicture}
\end{subfigure} \\
\normalsize{(b) w/ constraints} \\
    \caption{The distribution between the margin of predicted reward scores and the similarity of paired data (calculated by cosine similarity or length ratio) with or without constraint on the dialogue task.
    Each green point corresponds to a single data sample.
    The x-axis refers to the similarity calculated by the cosine similarity of sentence embedding or by the ratio of sentence length.
    The y-axis refers to the margin of the reward scores.}
    \label{fig:alpaca_reward_distribution}
\end{figure}
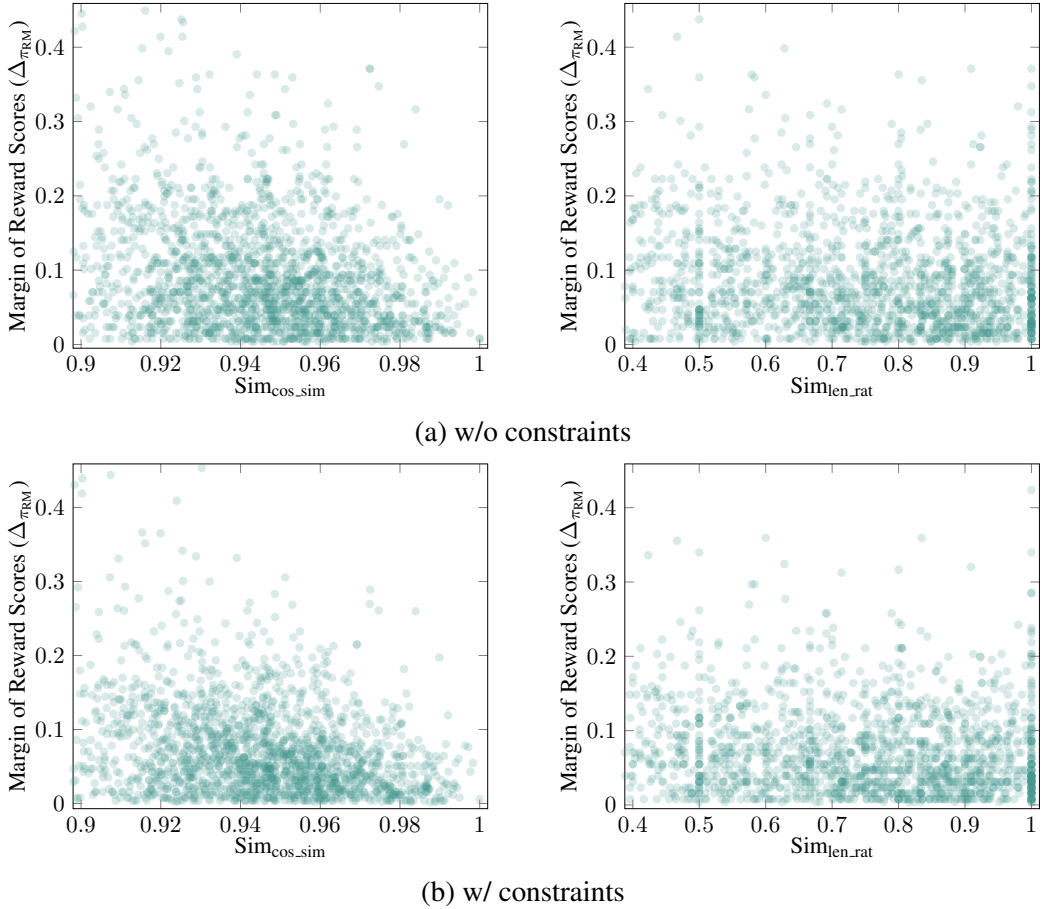

The performance of the PCRM on the dialogue and the summarization task are presented in Table \ref{tab:reward_accuracy}. We can see that the accuracy of the PCRM increases slightly, either by calculating similarity with cosine similarity or length ratio. As a comparison, we test the reward models trained with random and fixed constraints, whose performance is close to the vanilla one with some decrease. This result demonstrates the importance of the prior information. The appropriate value of the constraints also affects the performance because the training object of PCRM can be divided into two parts: achieving higher accuracy and controlling the distribution of the reward scores. The latter part may conflict with the former under extreme constraints (See Section \ref{subsubsec:value_constraints_on_PCRM}). 

To further explore the distribution of reward scores, we visualize the relationship between the margin of predicted reward scores and the similarity of the paired data on dialogue task, with or without constraints in Figure \ref{fig:alpaca_reward_distribution}.
The area with deeper colour in the figure represents more points gathering at that position. 
The points in Figure \ref{fig:alpaca_reward_distribution} (a) are more dispersed than the ones shown in Figure \ref{fig:alpaca_reward_distribution} (b) below, as the upper right part of the figures with constraints are more clean. 
From Figure \ref{fig:alpaca_reward_distribution} (b), we can observe that the margin of those points is limited to a lower level because of their higher similarity.
This observation confirms that our method can provide an effective constraint during the training of the reward model.  We can draw similar observations on the summarization task (See Figure \ref{fig:tldr_reward_distribution} in Appendix B).

\subsection{Results of Dialogue}

\begin{table}[!ht]
    \centering
    \begin{tabular}{l>{\centering\arraybackslash}p{5.2em}>{\centering\arraybackslash}p{5.2em}>{\centering\arraybackslash}p{5.2em}>{\centering\arraybackslash}p{5.2em}>{\centering\arraybackslash}p{5.2em}}
    \toprule
    \multirow{2}{*}{\textbf{Method}} & \multicolumn{2}{c}{\textbf{Dialogue}} & \multicolumn{3}{c}{\textbf{Summarization}} \\
    \cmidrule(r){2-3} \cmidrule(r){4-6}
     & PandaLM & GPT-4 Win & ROUGE-L & BARTScore & GPT-4 Win \\
    \midrule
    SFT & 65.75 & 55.77 & 22.93 & -6.31 & 41.83 \\
    \midrule
    RLHF & 72.73 & 60.13 & 25.10 & -5.17 & 70.53 \\
    PCRM-Random & 70.81 & 59.75 & 24.03 & -6.02 & 50.53 \\
    PCRM-Fixed & 69.34 & 60.71 & 24.87 & -6.09 & 52.21 \\
    PCRM-Length (Ours) & 73.52 & 62.30 & 22.76 & \textbf{-4.90} & 71.87 \\
    PCRM-Cosine (Ours) & \textbf{75.91} & \textbf{62.61} & \textbf{25.97} & -5.34 & \textbf{72.52} \\
    \bottomrule
\end{tabular}
    \caption{The performance of alignment with PCRM on downstream tasks.}
    \label{tab:downstream_tasks}
\end{table}

The results of our alignment with PCRM on downstream tasks, as detailed in Table \ref{tab:downstream_tasks}, show that PCRM significantly outperforms the vanilla reward model. Specifically, when utilizing cosine similarity, PCRM surpasses RLHF by +3.18 points on PandaLM and +2.48 points on the GPT-4 win rate in dialogue task.

The importance of meaningful constraints cannot be overstated. Our tests with PCRM using both random and fixed constraints revealed that these constraints were either marginally effective or even detrimental to alignment progress. Furthermore, we found that prior constraints enriched with additional information yielded better results. This is evident from comparing the effectiveness of length ratio and cosine similarity in calculating similarity; the superior performance of cosine similarity can be attributed to its ability to capture implicit prior semantic information of sentences, whereas the length ratio approach is more superficial and offers less valuable information.

\subsection{Results of Summarization}

Except for the dialogue task, we achieve similar results with our PCRM method on the summarization task as shown in Table \ref{tab:downstream_tasks}. Specifically, PCRM with cosine similarity outperforms RLHF by +0.87 points on ROUGE-L and +1.99 points on GPT-4 win rate. As to PCRM with length ratio, it outperforms RLHF by +0.27 points on BART Score and +1.34 points on GPT-4 win rate. It is obvious that there is a misalignment between ROUGE-L and BART Score- the models with high BART Score may not necessarily achieve high ROUGE-L scores. We attribute this phenomenon to the probable low correlation with human judgments, which is also reported in \newcite{wang2023learning}. 

\subsection{Analysis}

\subsubsection{Performance of the Variety Range and Numerical Size of Constraints on PCRM} \label{subsubsec:value_constraints_on_PCRM}

\begin{table}[t]
    \centering
    \begin{tabular}{>{\centering\arraybackslash}p{2em}>{\centering\arraybackslash}p{2em}>{\centering\arraybackslash}p{6.5em}>{\centering\arraybackslash}p{6.5em}}
    \toprule
    \multicolumn{1}{c}{$\boldsymbol{\beta_1}$} 
    & \multicolumn{1}{c}{$\boldsymbol{\beta_3}$} 
    & \renewcommand{\arraystretch}{.6}\begin{tabular}[x]{@{}c@{}}\textbf{Accuracy with}\\$\textrm{\textbf{Sim}}_{\textrm{\textbf{cos\_sim}}}(\cdot)$\end{tabular}\renewcommand{\arraystretch}{1} 
    & \renewcommand{\arraystretch}{.6}\begin{tabular}[x]{@{}c@{}}\textbf{Accuracy with}\\$\textrm{\textbf{Sim}}_{\textrm{\textbf{len\_rat}}}(\cdot)$\end{tabular}\renewcommand{\arraystretch}{1} \\
    \midrule
    10 & -3 & 54.79 & 55.87 \\
    10 & -5 & 55.73 & \textbf{55.87} \\
    10 & -7 & 53.92 & 55.33 \\
    10 & -9 & 52.26 & 51.60 \\
    20 & -13 & 53.73 & 55.33 \\
    20 & -15 & \textbf{56.53} & 55.73 \\
    20 & -17 & 54.80 & 55.47 \\
    20 & -19 & 55.20 & 55.20 \\
    30 & -25 & 54.80 & 55.60 \\
    40 & -35 & 55.60 & 55.07 \\
    \bottomrule
\end{tabular}

    \caption{The performance of PCRM with different constraints measured by accuracy. The range and the numerical size of the constraints are controlled with different pairs of $\beta_1$, $\beta_3$, and $\beta_2$ is fixed to 0.001 for computational stability.}
    \label{tab:beta_influence_on_reward_scores}
\end{table}

The variety range and numerical size of constraints are two key factors that control the maximum score margin and the strength of constraints.
Therefore, we conduct experiments to study the impact of the variety range and the numerical size of constraints.
Specifically, we explore with $\beta_2$ fixed to 0.001 in Equation \ref{eq:constraint_similarity} for computational stability, and different $\beta_1$, $\beta_3$, for variety range and numerical size on PCRM.
The results are summarized in Table \ref{tab:beta_influence_on_reward_scores}.
From the results, we can observe that the unsuitable constraints may hurt the performance.
We conjecture that the larger one may weaken the effect of distribution control, and the smaller one may conflict with the typical optimization.
Based on these experimental results, we find that the optimal $\beta_1$ and $\beta_3$ are 20 and -15 for the cosine similarity constraint; the optimal $\beta_1$ and $\beta_3$ are 10 and -5 for the length ratio constraint.
Note that $\beta_2$ prevents the occurrence of a similarity $\textrm{Sim}(\cdot)$ of 0, so we do not tune it and simply set a relatively small value.

\subsubsection{Integrating PCRM to DPO} \label{sec:extend_dpo}

\newcite{rafailov2024direct} proposes a direct preference optimization (DPO) method that bypasses the reward modeling step and directly optimizes an LLM using preference data.
However, this method still performs optimization based on ranking loss, which has the same limitation during training with preference data.
Therefore, we attempt to integrate the proposed PCRM into DPO, constraining the direct preference optimization.
Suppose we have the LLM trained by SFT $\pi^{\textrm{SFT}}_{\theta}$, and based on it, we fine-tune the model with DPO and preference data $(y_1,y_2,x)\sim\mathcal{D}$ where $y_1 \succ y_2$. \newcite{rafailov2024direct} points out that the reward model can be represented as follows with some mathematical transformation on the optimization objective of PPO:
\begin{equation}
    \pi^{\textrm{RM}}_\theta(y,x) = \beta\log\frac{\pi^{\textrm{RM}}_\theta(y|x)}{\pi^{\textrm{SFT}}_\theta(y|x)}+\beta\log Z(x)
\end{equation}
where $Z(x)=\Sigma_y \pi^{\textrm{SFT}}_\theta(y|x)\exp\left(\frac{1}{\beta}\pi^{\textrm{RM}}_\theta(y|x)\right)$. 
Bring the above equation into Equation \ref{eq:PCRM_prob} and \ref{eq:PCRM_loss}, we can obtain the loss of \textbf{P}rior \textbf{C}onstraints-based \textbf{DPO} (\textbf{PCDPO}):
\begin{eqnarray}
    \mathcal{L}_{\textrm{PCDPO}} = &-&\log P_{\pi^{\textrm{RM}}_\theta}\left(\Delta^* > \Delta_{\pi^{\textrm{RM}}_\theta}>0\right) \\
                    = &-&\log \sigma \left(\Delta^* - \Delta_{\pi^{\textrm{RM}}_\theta}\right) -\log \sigma \left(\Delta_{\pi^{\textrm{RM}}_\theta}\right) \\
                    = &-&\log \sigma \left(\Delta^*-\beta\log\frac{\pi^{\textrm{RM}}_\theta(y_1|x)}{\pi^{\textrm{SFT}}_\theta(y_1|x)}+\beta\log\frac{\pi^{\textrm{RM}}_\theta(y_2|x)}{\pi^{\textrm{SFT}}_\theta(y_2|x)}\right) \nonumber\\
                        &-& \log \sigma \left(\beta\log\frac{\pi^{\textrm{RM}}_\theta(y_1|x)}{\pi^{\textrm{SFT}}_\theta(y_1|x)}-\beta\log\frac{\pi^{\textrm{RM}}_\theta(y_2|x)}{\pi^{\textrm{SFT}}_\theta(y_2|x)}\right)
\end{eqnarray}

\begin{table}[t]
    \centering
    \begin{tabular}{lcc}
    \toprule
    \textbf{Method} & \textbf{PandaLM} & \textbf{GPT-4 Win} \\
    \midrule
    SFT & 65.75 & 55.77 \\
    \midrule
    DPO & 75.64 & 60.96 \\
    PCDPO-Length (Ours) & 77.02 & 61.11 \\
    PCDPO-Cosine (Ours) & \textbf{78.11} & \textbf{63.91} \\
    \bottomrule
\end{tabular}
    \caption{The experiment results of PCDPO.}
    \label{tab:PCDPO_alpaca}
\end{table}
With the theory above, we conduct experiments with the constraints on the alpaca dataset.
The experimental results are shown in Table \ref{tab:PCDPO_alpaca}.
From the results, we can find that PCRM can yield consistency improvements in DPO.
Notably, when armed with the cosine similarity constraint, we can obtain a +2.95 points improvement on the GPT-4 win rate.





\subsubsection{Inference with Different Temperature}

\begin{figure}[!ht]
    \centering
    \begin{tikzpicture}	
	\begin{axis}
		[
			ymajorgrids,
			xmajorgrids,
			grid style=dashed,
			anchor=north west,
			at={(0,0)},
			ymin=55, ymax=80, ytick={55,60,65,70,75,80},
			symbolic x coords={0.00,0.25,0.50,0.75,1.00},
			xtick=data,
			x tick label style={/pgf/number format/fixed,
				/pgf/number format/fixed zerofill,
				/pgf/number format/precision=1, scale=1.0},
			y tick label style={
				/pgf/number format/fixed, xshift=-0.5ex,
				/pgf/number format/fixed zerofill,
				/pgf/number format/precision=1, scale=1.0},
			ylabel=PandaLM,
			ylabel style={yshift=0,scale=1.0},
			xlabel=Sampling Temperature,
			xlabel style={yshift=0,scale=1.0},
            legend style={
                at={(0.5,1.2)},
                anchor=north,
                legend columns=2},
			legend cell align=left,
            enlarge x limits=0,
		]
		  
		\confidenceband [goodforeyesgreen, opacity=0.2, forget plot]{figures/data/alpaca/inference_with_diff_temperature.dat}{temp}{sft_max}{sft_min};
		\addplot [goodforeyesgreen,mark=triangle*,line width=.5pt] table [x=temp,y=sft_ave]{figures/data/alpaca/inference_with_diff_temperature.dat};
		\addlegendentry{SFT};
							            
		\confidenceband [lovelyred, opacity=0.2, forget plot]{figures/data/alpaca/inference_with_diff_temperature.dat}{temp}{ppo-baseline_max}{ppo-baseline_min};
		\addplot [lovelyred,mark=pentagon*,line width=.5pt] table [x=temp,y=ppo-baseline_ave]{figures/data/alpaca/inference_with_diff_temperature.dat};
		\addlegendentry{PPO-baseline};
		
		\confidenceband [supriseorange, opacity=0.2, forget plot]{figures/data/alpaca/inference_with_diff_temperature.dat}{temp}{ppo-lenrat_max}{ppo-lenrat_min};
		\addplot [supriseorange,mark=square*,line width=.5pt] table [x=temp,y=ppo-lenrat_ave]{figures/data/alpaca/inference_with_diff_temperature.dat};
		\addlegendentry{PPO-len\_rat};

		\confidenceband [sadblue, opacity=0.2, forget plot]{figures/data/alpaca/inference_with_diff_temperature.dat}{temp}{ppo-cossim_max}{ppo-cossim_min};
		\addplot [sadblue,mark=diamond*,line width=.5pt] table [x=temp,y=ppo-cossim_ave]{figures/data/alpaca/inference_with_diff_temperature.dat};
		\addlegendentry{PPO-cos\_sim};
	\end{axis}
\end{tikzpicture}
    \caption{PandaLM scores for different sampling temperatures using different methods. For each dialogue model, we conduct the generation three times and report the mean score of these generated responses.}
    \label{fig:inference_with_diff_temperature}
\end{figure}
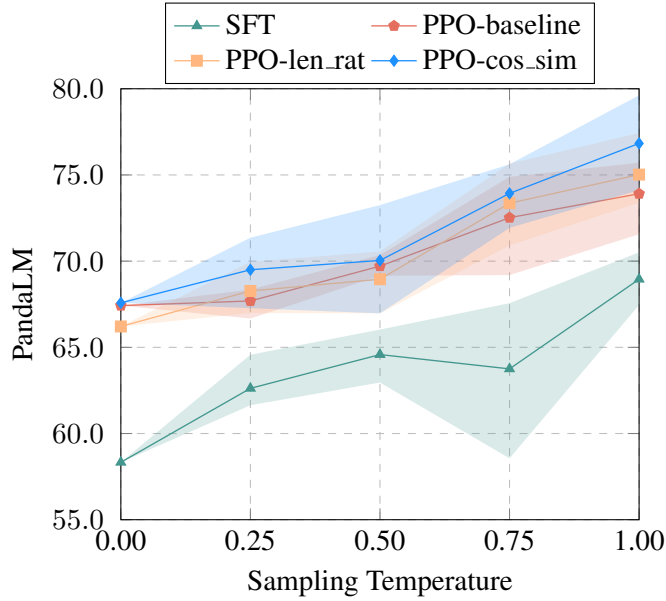

It is known to us that temperature influences the performance of inference, so we make inferences with different sampling temperatures on the dialogue task three times and calculate their mean in case of random variations while comparing different methods. Results are reported in Figure \ref{fig:inference_with_diff_temperature}.

The experiment results are consistent with the previous. Our method, PCRM with cosine similarity, outperforms the vanilla reward model no matter which sampling temperature is used, and the increase of PCRM with length ratio outperforms most of the temperatures. As analyzed before, we infer that the performance of the length ratio is limited due to the easy-to-learn prior information.

\section{Conclusions}

RLHF plays an important part in applications based on LLMs for their ability to align language models with human preferences. As a proxy for human preferences, the reward model makes a significant contribution.
We have introduced PCRM, a method of training reward models with prior constraints to better control the distribution of reward scores without sacrificing prediction accuracy. Instead of optimizing the vanilla reward model with the goal of maximizing the margin of scores, PCRM restricts the optimization process with the provided prior information and fits the distribution of reward scores for better alignment. Furthermore, this method is applicable to DPO, which treats the language model as a reward model and optimizes it directly. Through experiments on downstream tasks, we have validated the effectiveness of this method.

\section{Limitations}

Our results raise several questions that are beyond the scope of the present study: How can we determine the appropriate range of constraints for different tasks or datasets? For example, we explore hyperparameters in dialogue and summarization tasks, what about when it comes to a brand new task? Can the function of the constraints be replaced with any other prior information, and if so, will that be effective? Finally, it would be advantageous if the prior constraints could be learned automatically from data, rather than being manually set.

\section*{Acknowledgements}

This work was supported in part by the National Science Foundation of China (No.62276056), the Natural Science Foundation of Liaoning Province of China (2022-KF-16-01), the Fundamental Research Funds for the Central Universities (Nos. N2216016 and N2316002), the Yunnan Fundamental Research Projects (No. 202401BC070021), and the Program of Introducing Talents of Discipline to Universities, Plan 111 (No.B16009).


\clearpage

\bibliographystyle{ccl}
\bibliography{reference}








\clearpage

\appendix

\section*{Appendix A. Calculating Accuracy of the Reward Scores} \label{appendix:calculating_accuracy_of_the_reward_scores}

Suppose we have a test set for human preference $\left(x,y_1,y_2\right)\sim\mathcal{D}_\textrm{test}$, in which $y_1$ is preferred than $y_2$ with the same $x$ by human, and the corresponding scores predicted by the reward model are $\pi^{\textrm{RM}}_\theta(y1,x)$, $\pi^{\textrm{RM}}_\theta(y2,x)$. The accuracy of the scores is defined as:
\begin{eqnarray}
\textrm{Acc}(\pi^{\textrm{RM}}_\theta, \mathcal{D}_\textrm{test})=
\frac{\textrm{Count}_{\left(x,y_1,y_2\right)\sim\mathcal{D}_\textrm{test}}\left(\pi^{\textrm{RM}}_\theta(y1,x)>\pi^{\textrm{RM}}_\theta(y2,x)\right)}{\textrm{Count}\left(\left(x,y_1,y_2\right)\in\mathcal{D}_\textrm{test}\right)}
\end{eqnarray}
where $\textrm{Count}(\cdot)$ denotes the total number of the samples that meet the condition.

\section*{Appendix B. Distribution of Reward Scores for Summarization Task}\label{appendix:tldr_reward_distribution_analysis}

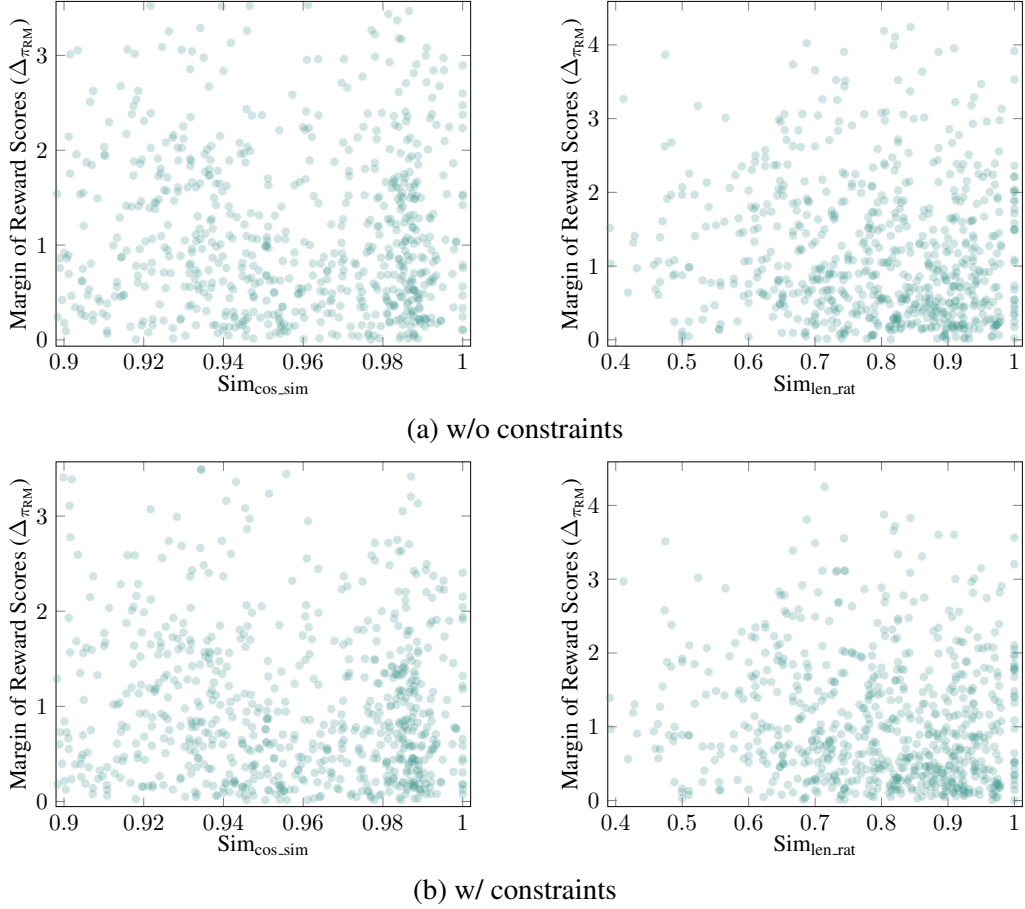
\begin{figure}[!ht]
    \centering
    \begin{subfigure}[b]{.45\textwidth}
	\centering
	\begin{tikzpicture}[domain=0:1, scale=0.8]
		\begin{axis}[xlabel={$\textrm{Sim}_{\textrm{cos\_sim}}$}, ylabel={\normalsize{Margin of Reward Scores ($\Delta_{\pi_{\textrm{RM}}}$})},
			y label style={yshift=-.5em}, x label style={yshift=1ex}, enlarge x limits=0.02, enlarge y limits=0.02, xmin=.9, ymax=3.5]
			\addplot [scatter, only marks,  scatter src=explicit symbolic,
				scatter/classes={
					r={mark=*,goodforeyesgreen, fill opacity=0.25, draw opacity=0},
					w={mark=diamond*,draw=black!30,fill=black!30,fill opacity=0.8, draw opacity=0}
			}]
			table [x=sim, y=delta,meta=label]{"figures/data/tldr/baseline-epoch1-delta-sim_cos_sim.dat"};
		\end{axis}
	\end{tikzpicture}
\end{subfigure}
\begin{subfigure}[b]{.45\textwidth}
	\centering
	\begin{tikzpicture}[domain=0:1,scale=0.8]
		\begin{axis}[xlabel={$\textrm{Sim}_{\textrm{len\_rat}}$}, ylabel={\normalsize{Margin of Reward Scores ($\Delta_{\pi_{\textrm{RM}}}$})},
			y label style={yshift=-.5em}, x label style={yshift=1ex}, enlarge x limits=0.02, enlarge y limits=0.02, xmin=.4, ymax=4.5]
			\addplot [scatter, only marks,  scatter src=explicit symbolic,
				scatter/classes={
					r={mark=*,goodforeyesgreen, fill opacity=0.25, draw opacity=0},
					w={mark=diamond*,draw=black!30,fill=black!30,fill opacity=0.8, draw opacity=0}
			}]
			table [x=sim, y=delta,meta=label]{"figures/data/tldr/baseline-epoch1-delta-sim_len_rat.dat"};
		\end{axis}
	\end{tikzpicture}
\end{subfigure} \\
\normalsize{(a) w/o constraints} \\
\vspace{0.2cm}
\begin{subfigure}[b]{.45\textwidth}
	\centering
	\begin{tikzpicture}[domain=0:1,scale=0.8]
		\begin{axis}[xlabel={$\textrm{Sim}_{\textrm{cos\_sim}}$}, ylabel={\normalsize{Margin of Reward Scores ($\Delta_{\pi_{\textrm{RM}}}$})},
			y label style={yshift=-.5em}, x label style={yshift=1ex}, enlarge x limits=0.02, enlarge y limits=0.02, xmin=.9, ymax=3.5]
			\addplot [scatter, only marks,  scatter src=explicit symbolic,
				scatter/classes={
					r={mark=*,goodforeyesgreen, fill opacity=0.25, draw opacity=0},
					w={mark=diamond*,draw=black!30,fill=black!30,fill opacity=0.8, draw opacity=0}
			}]
			table [x=sim, y=delta,meta=label]{"figures/data/tldr/cos_sim-beta1_20_beta3_-15-delta-sim_cos_sim.dat"};
		\end{axis}
	\end{tikzpicture}
\end{subfigure}
\begin{subfigure}[b]{.45\textwidth}
	\centering
	\begin{tikzpicture}[domain=0:1,scale=0.8]
		\begin{axis}[xlabel={$\textrm{Sim}_{\textrm{len\_rat}}$}, ylabel={\normalsize{Margin of Reward Scores ($\Delta_{\pi_{\textrm{RM}}}$})},
			y label style={yshift=-.5em}, x label style={yshift=1ex}, enlarge x limits=0.02, enlarge y limits=0.02, xmin=0.4, ymax=4.5]
			\addplot [scatter, only marks,  scatter src=explicit symbolic,
				scatter/classes={
					r={mark=*,goodforeyesgreen, fill opacity=0.25, draw opacity=0},
					w={mark=diamond*,draw=black!30,fill=black!30,fill opacity=0.8, draw opacity=0}
			}]
			table [x=sim, y=delta,meta=label]{"figures/data/tldr/len_rat-beta1_10_beta3_-5-delta-sim_len_rat.dat"};
		\end{axis}
	\end{tikzpicture}
\end{subfigure} \\
\normalsize{(b) w/ constraints} \\
    \caption{The distribution between the margin of predicted reward scores and the similarity of paired data (calculated by cosine similarity or length ratio) with or without constraint on the summarization task.}
    \label{fig:tldr_reward_distribution}
\end{figure}

\end{document}